\documentclass[10pt,twocolumn,letterpaper]{article}

\usepackage{iccv}
\usepackage{times}
\usepackage{epsfig}
\usepackage{graphicx}
\usepackage{amsmath}
\usepackage{amssymb}

\usepackage[breaklinks=true,bookmarks=false]{hyperref}
\usepackage{multirow} 
\usepackage{pifont} 

\iccvfinalcopy 

\ificcvfinal\fi

\begin{document}

\title{MSAF: Multimodal Split Attention Fusion}

\author{
    Lang Su,\textsuperscript{\rm 1}\thanks{Equal Contribution.}\;
    Chuqing Hu,\textsuperscript{\rm 1}\footnotemark[1]\;
    Guofa Li,\textsuperscript{\rm 2} 
    Dongpu Cao\textsuperscript{\rm 1} 
    \\
    \textsuperscript{\rm 1} University of Waterloo, \textsuperscript{\rm 2} Shenzhen University\\
    \tt\small\{l26su, cq2hu, dongpu\}@uwaterloo.ca, guofali@szu.edu.cn
}

\maketitle
\ificcvfinal\thispagestyle{empty}\fi

\begin{abstract}
Multimodal learning mimics the reasoning process of the human multi-sensory system, which is used to perceive the surrounding world. While making a prediction, the human brain tends to relate crucial cues from multiple sources of information. In this work, we propose a novel lightweight multimodal fusion module that learns to emphasize more contributive features across all modalities. Specifically, the proposed Multimodal Split Attention Fusion (MSAF) module splits each modality into channel-wise equal feature blocks and creates a joint representation that is used to generate soft attention for each channel across the feature blocks. Further, the MSAF module is designed to be compatible with features of various spatial dimensions and sequence lengths, suitable for both CNNs and RNNs. Thus, MSAF can be easily added to fuse features of any unimodal networks and utilize existing pretrained unimodal model weights. To demonstrate the effectiveness of our fusion module, we design three multimodal networks with MSAF for emotion recognition, sentiment analysis, and action recognition tasks. Our approach achieves competitive results in each task and outperforms other application-specific networks and multimodal fusion benchmarks. 
\end{abstract}

\section{Introduction}
Multimodal learning has been explored in numerous machine learning applications such as audio-visual speech recognition \cite{neti2000audio}, action recognition \cite{baccouche2011sequential}, and video question answering \cite{lei2018tvqa}, where each modality contains useful information from a different perspective. Although these tasks can benefit from the complementary relationship in multimodal data, different modalities are represented in diverse fashions, making it challenging to grasp their complex correlations. 

Studies in multimodal machine learning are mainly categorized into three fusion strategies: early fusion, intermediate fusion, and late fusion. Early fusion explicitly exploits the cross-modal correlation by joining the representation of the features from each modality at the feature level, which is then used to predict the final outcome. The fusion is typically operated after the feature extractor for each modality, where techniques such as Compact Bilinear Pooling (CBP) \cite{fukui2016multimodal, nguyen2018deep} and Canonical Correlation Analysis (CCA) \cite{liu2019multimodal, mittal2020m3er} are used to exploit the covariation between modalities. Unfortunately, modalities usually have different natures causing unaligned spatial and temporal dimensions. This creates obstacles in capturing the latent interrelationships in the low-level feature space \cite{baltruvsaitis2018multimodal}. On the other hand, late fusion fuses the decision from each modality into a final decision using a simple mechanism such as voting \cite{morvant2014majority} and averaging \cite{shutova-etal-2016-black}. Since little training is required, a multimodal system can be promptly deployed by utilizing pretrained unimodal weights, unlike early fusion methods. However, decision-level fusion neglects the crossmodal correlation between the low-level features in modalities, resulting in limited improvement compared to the unimodal models. The intermediate fusion method joins features in the middle of the network, where some feature processing is done for the raw features from the feature extractors. Recent intermediate multimodal fusion networks \cite{joze2020mmtm, vielzeuf2018centralnet, hu2019dense} exploit the modality-wise relationships at different stages of the network, which has shown impressive results. However, there are still a limited number of works that can effectively capture cross-modal dynamics in an efficient way by using pretrained weights while introducing minimal parameters. 

To overcome the aforementioned shortcomings of intermediate fusion methods, we propose a lightweight fusion module, MSAF, taking inspiration from the split-attention block in ResNeSt \cite{zhang2020resnest}. The split-attention mechanism explores cross-channel relationships by dividing the feature-map into several groups and applying attention across the groups based on the global contextual information. We extend split-attention for multimodal applications in the proposed MSAF module to explore inter- and intra-modality relationships while maintaining a compact multimodal context. The MSAF module splits the features of each modality channel-wise into equal-sized feature blocks, which are globally summarized by a channel descriptor. The descriptor then learns to emphasize the important feature blocks by generating attention values. Subsequently, the enhanced feature blocks are rejoined for each modality, resulting in an optimized feature space with an understanding of the multimodal context. Our MSAF module is compatible with features of any shape as it operates only on the channel dimension. Thus, MSAF can be added between layers of any CNN or RNN architecture. Furthermore, we boost performance on sequential features by splitting modalities time-wise and applying an MSAF module to each time segment. This allows emphasis of different feature blocks in each time segment. To our knowledge, our work is the first independent fusion module that can be used in both CNN- and RNN-based multimodal learning applications.

We comprehensively evaluate the effectiveness of MSAF in three multimodal learning tasks, namely audiovisual emotion recognition, sentiment analysis, and action recognition. We design a neural network with integrated MSAF modules for each task to demonstrate the ease of applying MSAF to existing unimodal configurations. Our experiments show that MSAF-powered networks outperform other fusion methods and state-of-the-art models designed for each application. Empirically, we observe that MSAF achieves better results while using a similar number of parameters as simple late fusion methods. Our module learns to pinpoint the important features regardless of the modality's complexity. 

In summary, our work provides the following contributions: 1) MSAF -- A novel lightweight multimodal fusion module for CNN and RNN networks that effectively fuses intermediate and high-level modality features to leverage the advantages of each modality. 2) We design three multimodal fusion networks corresponding to three applications: emotion recognition, sentiment analysis, and action recognition. Our experiments demonstrate the capabilities of MSAF through competitive results in all applications while utilizing fewer parameters.

\section{Related Work}
\paragraph{Early Fusion.}
The majority of works in early fusion integrate features immediately after they are extracted from each modality, whereas occasionally studies perform fusion at the input level, such as \cite{Morales2016DeepCA}. A simple solution for early fusion is feature concatenation after they are transformed to the same length, followed by fully connected layers. Many early fusion works use CCA to exploit cross-modality correlations. \cite{sargin2006multimodal} applies CCA to improve the performance in speaker identification using visual and audio modalities. \cite{andrew2013deep} proposes deep CCA to learn complex nonlinear transformations between modalities, which inspired multimodal applications such as \cite{liu2019multimodal}. Bilinear pooling is another early fusion method that fuses modalities by calculating their outer product. However, the generated high dimensional feature vectors are very computationally expensive for subsequent analysis. Compact bilinear pooling \cite{gao2016compact} significantly mitigates the curse of dimensionality problem \cite{hu2019dense} through a novel kernelized analysis while keeping the same discriminative power as the full bilinear representation. 

\paragraph{Late Fusion.}
Late fusion merges the decision values from each unimodal model into an unified decision using fusion mechanisms such as averaging \cite{shutova-etal-2016-black}, voting \cite{morvant2014majority} and weighted sum \cite{vora2014improved}. In contrast to early fusion, late fusion embraces the end-to-end learning between each modality and the given task. It allows for more flexibility as it can still train or make predictions when one or more modalities are missing. Nevertheless, late fusion lacks the exploration of lower-level correlations between the modalities. Therefore, when it comes to a disagreement between modalities, a simple mechanism acting only on decisions might be too simplified. There are also more complex late fusion approaches that exploit modality-wise synergies. For example, \cite{liu2018learn} proposes a multiplicative combination layer that promotes the training of strong modalities per sample and tolerates mistakes made by other modalities. 

\paragraph{Intermediate Fusion.}
Intermediate fusion exploits feature correlations after some level of processing, therefore the fusion takes place in the middle between the feature extractor and the decision layer. For instance, \cite{williams2018dnn} applies principle component analysis on the extracted features for each modality, and further processes them respectively before feature concatenation. Recent works continue to improve modality feature alignment to give stronger joint features. CentralNet \cite{vielzeuf2018centralnet} coordinates features of each modality by performing a weighted sum of modalities in a central branch at different levels of the network. EmbraceNet \cite{choi2019embracenet} prevents dependency on data of specific modalities and increases robustness to missing data through learning crossmodal correlations by combining selected features from each modality using a multinomial distribution. \cite{joze2020mmtm} utilizes the squeeze and excitation module from SENet \cite{hu2018squeeze} to enable slow modality fusion by channel-wise feature recalibration at different stages of the network. Our work aims to effectively fuse features of modalities while maintaining efficiency. We adopt the split attention \cite{zhang2020resnest} concept for multimodal fusion where modalities are broken down into feature map groups with hidden complement relationships. 

\section{Proposed Method}
\begin{figure}[t]
\centering
\includegraphics[width=0.9\columnwidth]{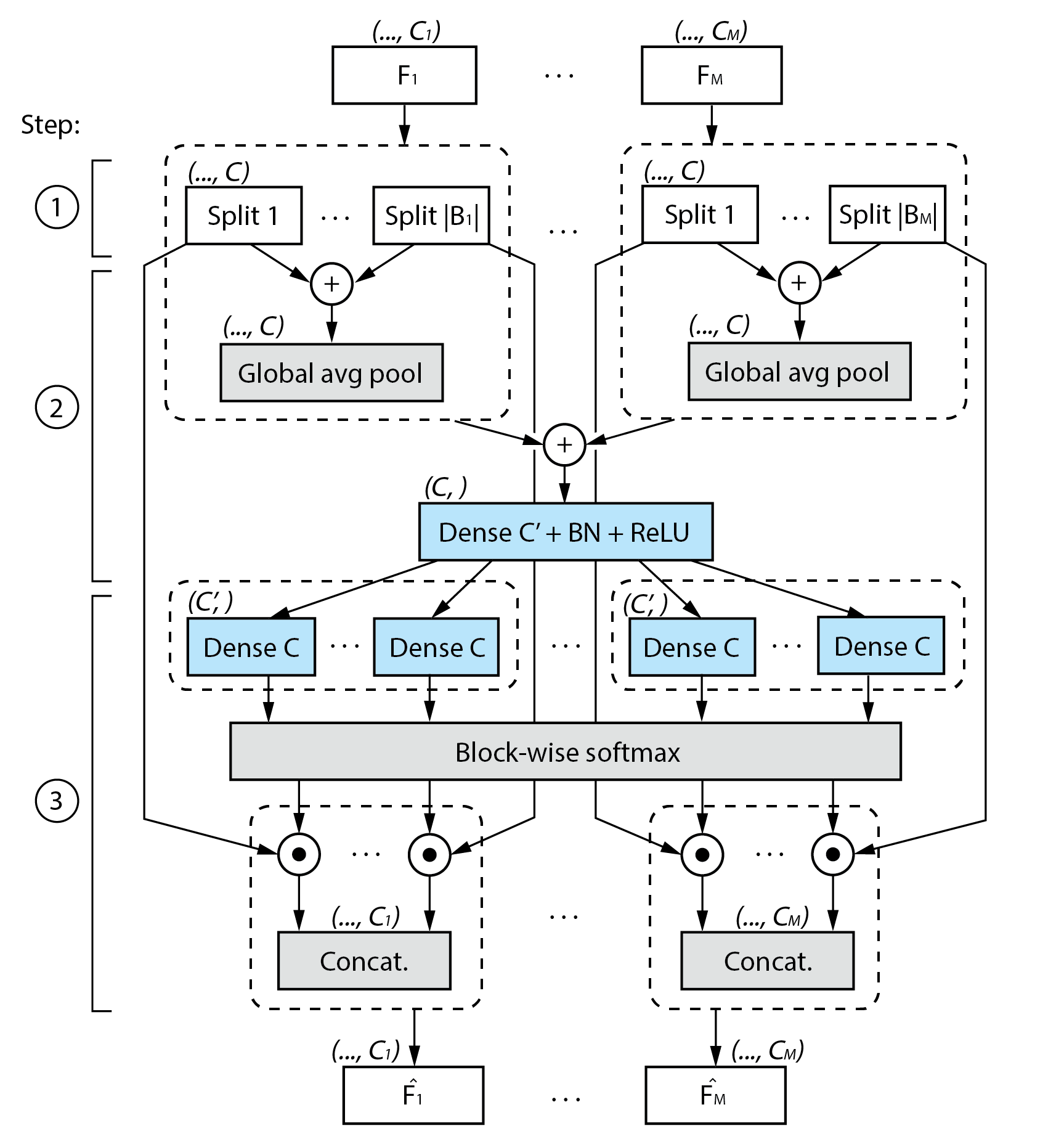} 
\caption{Breakdown of the MSAF module with steps, split, join and highlight, numbered on the left.}
\label{fig2}
\end{figure}

We first formulate the multimodal fusion problem in an MSAF module. Let $M$ be the number of modalities and the feature map of modality $m \in \{1, 2, \cdots, M\}$ be $F_m \in {\rm I\!R}^{N_1 \times N_2 \times \cdots \times N_K \times C_m}$. Here, $K$ is the number of spatial dimensions of modality $m$ and $C_m$ is the number of channels in modality $m$. Generally, an MSAF module takes the feature maps $\{F_1, \cdots, F_M\}$ and generates optimized feature maps $\{\hat{F_1}, \cdots, \hat{F_M}\}$ activated by the corresponding per channel block-wise attention. An MSAF module consists of three operations: 1) split, 2) join, and 3) highlight, which are summarized in Figure \ref{fig2}. We explicate the three steps below.

\paragraph{Split.}
We start by splitting each feature map channel-wise into equal-channel feature blocks where the number of channels in each block is $C$. We denote the set of the feature blocks that belong to modality $m$ as $B_m$, where $|B_m| = \lceil {C_m / C }\rceil , m \in \{1, \cdots, M\}$, $ B_m^i $ being the $i$th feature block in $B_m$, $i \in \{1, \cdots, |B_m|\}$. When $C_m$ is not a multiple of $C$, the last block is padded with zeros in the missing channels. 

\paragraph{Join.}
The join operation is a crucial step that learns the multimodal global context which is used to generate per channel block-wise attention. We join the blocks that belong to modality $m$ into a shared representation $D_m$, by calculating the element-wise sum $S_m$ over $B_m$, followed by global average pooling on the spatial dimensions: 
\begin{equation}
    D_m (c) = {1 \over \prod_{i=1}^{K} N_i} \sum_{(n_1, \cdots, n_K)} S_m(n_1, n_2, \cdots, n_K, c)
\end{equation}
Each channel descriptor is now a feature vector of the common length $C$ that summarizes the feature blocks within a modality. To obtain multimodal contextual information, we calculate the element-wise sum of the per modality descriptors $\{D_1, \cdots, D_M\}$ to form a multimodal channel descriptor $G$.
We capture the channel-wise dependencies by a fully connected layer with a reduction factor $r$ followed by a batch normalization layer and a ReLU activation function. The transformation maps $G$ to the joint representation $Z \in {\rm I\!R}^{C^{\prime}}$, $C'=\lfloor C/r \rfloor$ which helps with generalization for complex models. 
\begin{equation}
    Z = W_Z G + b_Z
\end{equation}
where $W_Z \in {\rm I\!R}^{C^{\prime}\times C}, b_Z \in {\rm I\!R}^{C^{\prime}}$. As advised in \cite{joze2020mmtm} and evident in our experiments, a reduction factor of 4 is ideal for two modalities. As the number of modalities increase, we recommend reducing the reduction factor to accommodate mores features in the joint representation. For example, when fusing 3 modalities, a reduction factor of 2 was optimal in our results for sentiment analysis.

\paragraph{Highlight.}
The multimodal channel descriptor contains generalized but rich knowledge of the global context. In this step, for a block $B_m^i$, we generate the corresponding logits $U_m^i$ by applying a linear transformation on $Z$ and obtain the block-wise attention weights $A_m^i$ using the softmax activation. 
\begin{equation}
    U_m^i = W_m^i Z + b_m^i
\end{equation}
\begin{equation}
    A_m^i = \frac{exp(U_m^i)}{\sum_{k}^{M} \sum_j^{|B_k|} exp(U^j_k)}\
    \label{atteneq}
\end{equation}
where $W_m^i \in {\rm I\!R}^{C \times C^{\prime}}$ and $b_m^i \in {\rm I\!R}^{C}$ are weights and bias of the corresponding fully connected layer.
Since soft attention values are dependent on the total number of feature blocks, features may be over-suppressed. The effect is more apparent in complex tasks which results in insufficient information for accurate predictions. Thus, we present a hyperparameter $\lambda \in [0, 1]$ that controls the suppression power of MSAF. Intuitively, $\lambda$ can be understood as a regularizer for the lowest attention of a split. We obtain an optimized feature block $\hat{B_m^i}$ using attention signals $A_m^i$ and $\lambda$:
\begin{equation}
    \hat{B_m^i} = [\lambda + (1 - \lambda) \times A_m^i] \odot B_m^i
\end{equation}
Finally, the feature blocks belonging to modality $m$ are merged by channel-wise concatenation to produce $\hat{F_m}$.
\begin{equation}
    \hat{F_m} = [ \hat{B_m^1}, \hat{B_m^2}, \cdots, \hat{B_m}^{|B_m|} ]
\end{equation}

\subsection{BlockDropout}
To lessen the dependencies on certain strong feature blocks and ease overfitting, we propose a dropout method for the feature blocks called BlockDropout. BlockDropout generates a binary mask that randomly drops feature blocks from the set of all feature blocks from each modality $B$, and applies the same mask on the block’s attention. Let the dropout probability $p \in [0, 1)$, we draw $|B|$ samples from a Bernoulli distribution with the probability of success $(1 - p)$, resulting in a binary mask for dropping out the feature blocks. Subsequently, the mask is scaled by $\frac{1}{1 - p}$ and is applied to the generated attention vectors. This is not to be confused with DropBlock \cite{NEURIPS2018_7edcfb2d} which is used in ResNeSt to regularize convolutional layers by  randomly masking out local block regions in the feature map. Whereas BlockDropout is applied to feature blocks after the first step of MSAF which are split in the channel dimension.

\subsection{Enhancing MSAF for RNNs}
As features pass through a CNN, the number of channels gradually increases while other dimensions shrink from convolving filters and pooling layers. In a multilayer RNN, the length of the feature sequence remains the same between layers. When applying split attention to features of longer sequence length, as is the case for RNNs, the original MSAF module does not have the flexibility to adjust the attention vector of each block over the sequence. For example, in audiovisual emotion recognition, there may be a segment where the audio modality should be highlighted over the video modality and vice versa in another segment of the same sequence.

\begin{figure}[t]
\centering
\includegraphics[width=0.9\columnwidth]{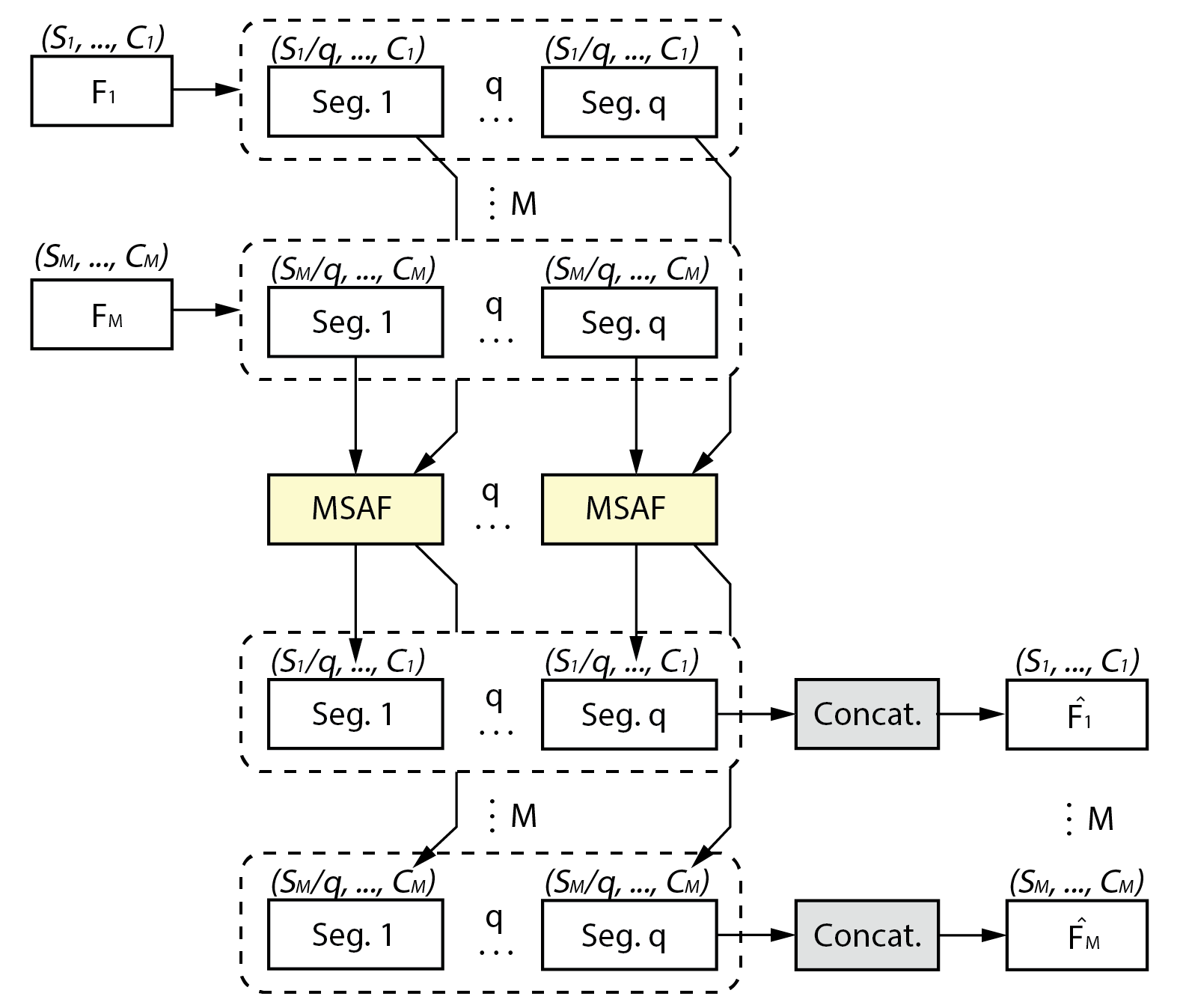} 
\caption{Enhancing the MSAF module for RNNs. For easy visualization, the sequence lengths are evenly divisible by $q$.}
\label{fig3}
\end{figure}

We achieve this by splitting each modality into $q$ segments. We define the sequence length of a modality as $S$ then the length of each segment for modality $m$ is $\lfloor S_m/q \rfloor$. The modalities of each segment are passed into separate MSAF blocks. Optimally the sequence lengths are the same or evenly divisible by $q$. Otherwise, the extra segment is combined with the second last segment. This process is visualized in Figure \ref{fig3}.

\section{Applications}
In this section, we apply the MSAF module to fuse unimodal networks in three applications. We describe each unimodal network and our configuration for the MSAF modules.

\subsection{Emotion Recognition}
\begin{figure}[ht]
\centering
\includegraphics[width=0.65\columnwidth]{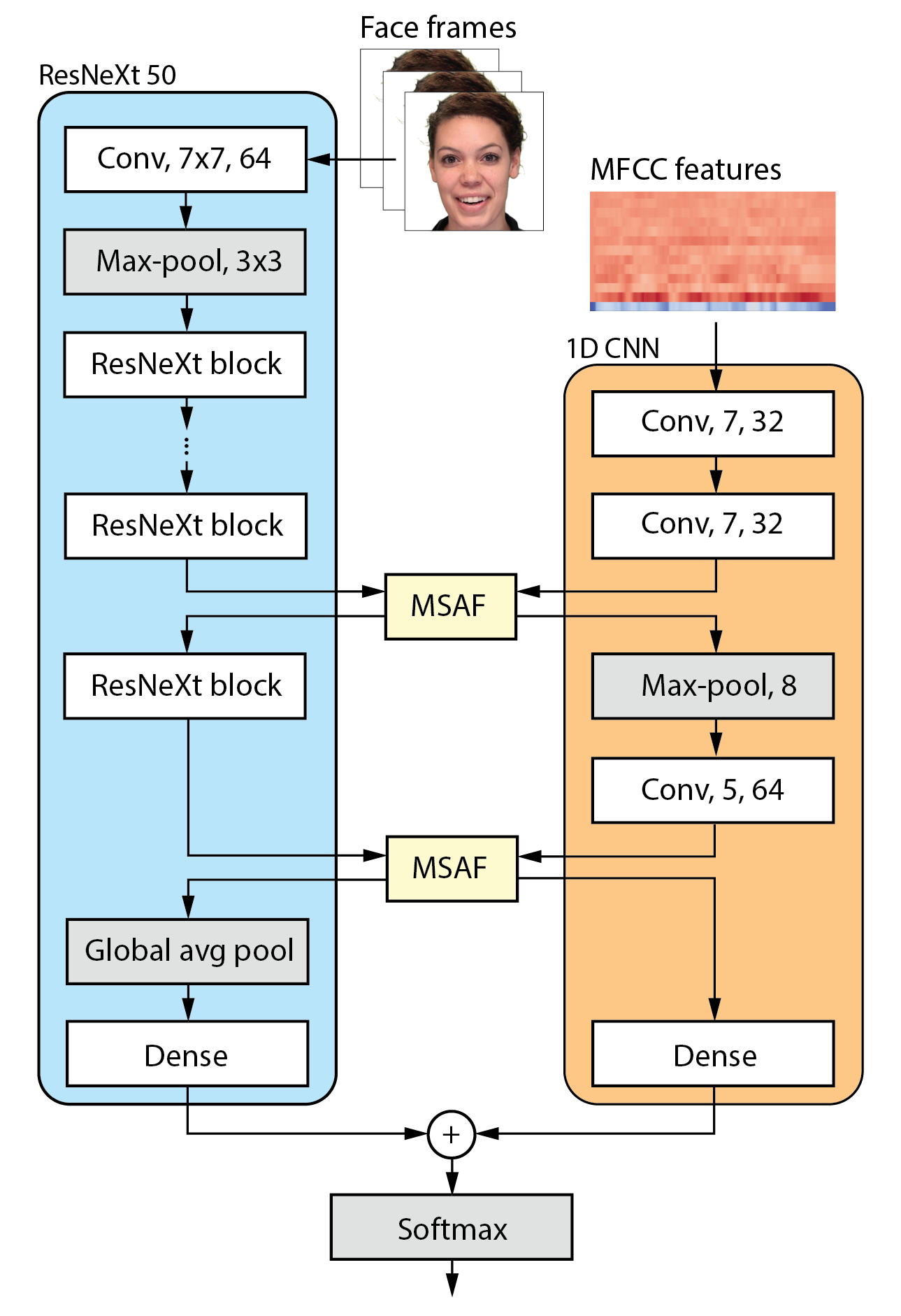} 
\caption{Proposed architecture for emotion recognition}
\label{fig4}
\end{figure}

Multimodal emotion recognition (MER) is a classification task that categorizes human emotions using multiple interacting signals. Although numerous works have utilized more complex modalities such as EEG \cite{zheng2014multimodal} and body gesture \cite{de2006towards}, video and audio remain as dominant modalities used for this task. Thus, we design a multimodal network that fuses a 3D CNN for video and a 1D CNN for audio using MSAF. Video data has dependencies in both spatial and temporal dimensions, therefore requires a network with 3D kernels to learn both the facial expression and its movement. Considering both network performance and training efficiency, we choose the 3D ResNeXt50 network \cite{xie2016aggregated} as suggested by \cite{hara2017can} with cardinality set to 32. For the audio modality, recent works \cite{neverova2015moddrop, wang2020speech} have demonstrated the effectiveness of deep learning based methods built on Mel-frequency cepstral coefficients (MFCC) features. We design a simple 1D CNN for the MFCC features and fuse the two modalities via two MSAF modules as shown in Figure \ref{fig4}. We configured the two MSAF modules with 16 and 32 channels per block respectively and BlockDropout with $p=0.2$. Finally, we sum the logits of both networks followed by a softmax function.

\subsection{Sentiment Analysis}
Sentiment analysis is the use of Natural Language Processing (NLP) to interpret and classify people’s opinions from text. In recent years, the multimodal nature of human language has led to the incorporation of other modalities such as visual and audio data in NLP tasks. Similar to works such as \cite{sun2019learning,hazarika2020misa}, we apply our module on audio, visual and text modalities. Our architecture uses two LSTM layers for each modality with an MSAF module in between to fuse features from the first LSTM layer. We use 32, 64 and 128 hidden units for both LSTM layers in the visual, audio and text modality respectively. For the MSAF module, each modality is split into 5 segments sequence-wise ($q=5$) and each segment is passed into a separate MSAF block with 16 channels per feature block and BlockDropout with $p=0.2$. Lastly, we concatenate the final output from each LSTM and pass that to a fully connected layer to generate the sentiment value.

\subsection{Action Recognition}
\begin{figure}[t]
\centering
\includegraphics[width=0.9\columnwidth]{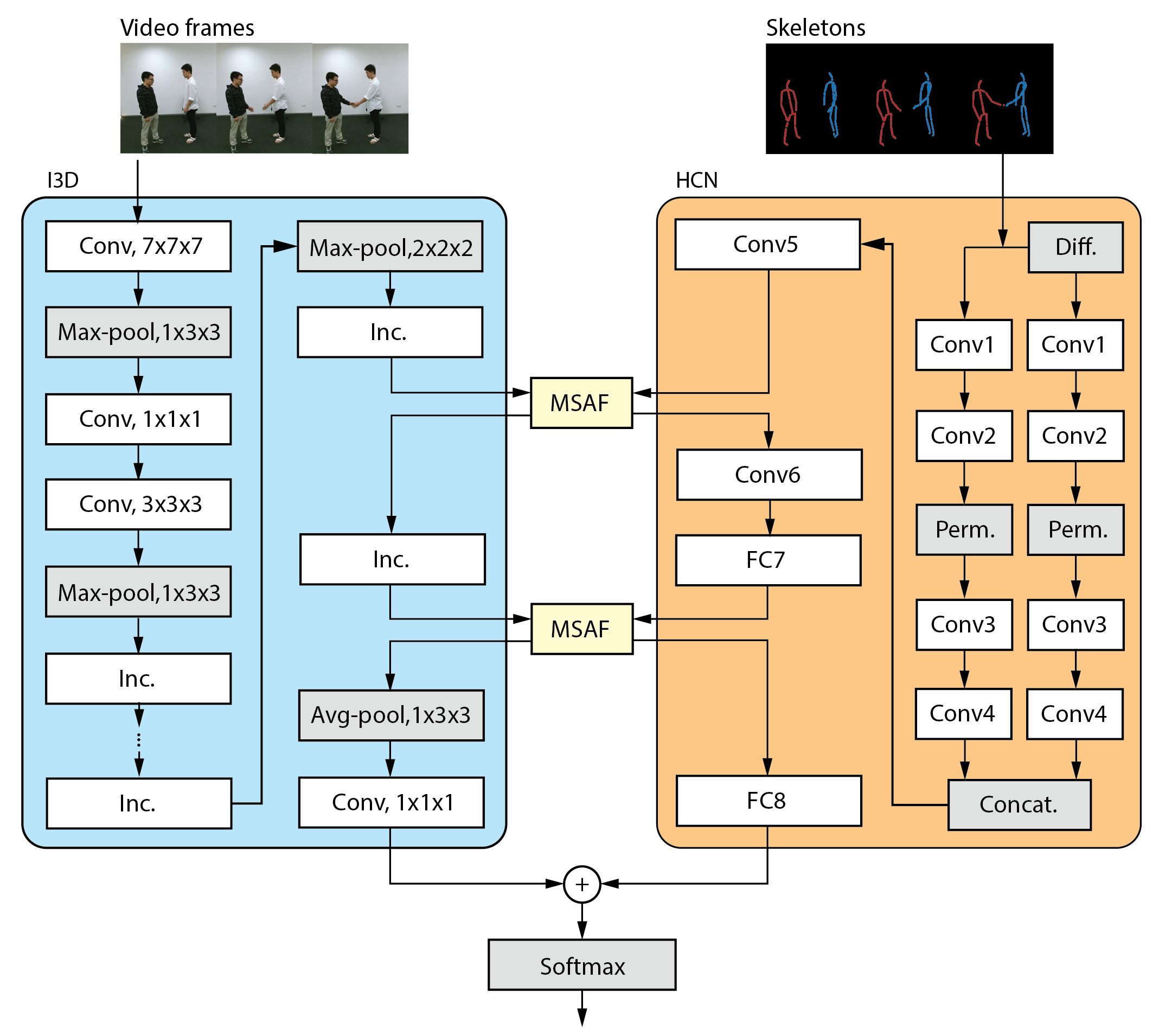} 
\caption{Proposed architecture for action recognition. ``Inc." denotes an inception module from \cite{carreira2017quo}}
\label{fig5}
\end{figure}

With the development of depth cameras, depth and skeleton data became crucial modalities in the action recognition task along with RGB videos. Multiple works such as \cite{joze2020mmtm, li2020sgm, liu2018recognizing} have achieved competitive performance using RGB videos associated with skeleton sequences. We follow \cite{joze2020mmtm} which utilizes I3D \cite{carreira2017quo} for the video data, and HCN \cite{ijcai2018-0109} for the skeleton stream. As illustrated in Figure \ref{fig5}, we deploy two MSAF modules, one at an intermediate-level in both networks, and the other one for high-level feature recalibration. The HCN framework proposes two strategies to be scalable to multi-person scenarios. The first type stacks the joints from all persons and feeds it as the input of the network in an early fusion style. The second type adapts late fusion that passes the inputs of multiple persons through the same subnetwork, whose Conv6 channel-wise concatenates or element-wise maximizes the group of features of persons. The latter generalizes better to various numbers of persons than the other, which needs a predefined maximum number of persons. \cite{joze2020mmtm} follows the multi-person late fusion strategy and utilizes their first fusion module on one of the two persons universally. We take a different approach by considering all available individuals in a sample because either can send important signals during a multi-person interaction. Our first MSAF module has 64 channels per block and is inserted between the second last Inception layer in I3D and the Conv5 outputs of each person. The second MSAF has 256 channels per block and is positioned between the last Inception layer in I3D and the FC7 layer in HCN. A suppression power of $\lambda=0.5$ is used for both modules. Finally, we average the logits of both networks followed by a softmax function. 

\section{Experiments}
In this section, we discuss our dataset choice, data preprocessing and training details for each application. Further, we evaluate our method and compare with other state-of-the-art works. Validation set accuracy was used to select the optimal hyperparameters for benchmarks and our proposed method in our tables. To verify the effectiveness of MSAF and the proposed hyperparameters, we conduct an ablation study for each task and analyze the module through its complexity, computation cost and visualization of attention signals. All of our experiments were conducted using a single Nvidia 2080 Ti GPU in Ubuntu 20.04 with Python 3.6 and PyTorch 1.7.1. To ensure the reproducibility of our results, our code is made publicly available on GitHub.

\subsection{Emotion Recognition}
\paragraph{Data Preparation.}
There are many emotion recognition datasets that contain both facial expression and audio signals including \cite{Livingstone2018TheRA, li2020spontaneous}. We chose the Ryerson Audio-Visual Database of Emotional Speech and Song (RAVDESS) \cite{Livingstone2018TheRA} dataset due to its high quality in both video and audio recording and the sufficient number of video clips. RAVDESS contains 1440 videos of short speech from 24 actors (12 males, 12 females), performed under the emotion they are told to act upon. Eight emotions are included in the dataset: neutral, calm, happy, sad, angry, fearful, disgust and surprised. We extracted 30 consecutive images from each video. For each image, we used the 2D facial landmarks provided to crop the face area and then resized to (224, 224). Random crop, horizontal flip, and normalization were used for data augmentation. For the audio modality, since the first 0.5 seconds usually contains no sound, we cropped the first 0.5 seconds and took the next 2.45 seconds for consistency. As suggested by \cite{jin2015speech}, we extracted the first 13 MFCC features for each cropped audio clip. We proceed with a 6 fold cross-validation based on the actors for the RAVDESS dataset. We split the 24 actors in a 5:1 ratio for the training and testing sets. Since the gender of the actors is indicated by even or odd actor IDs, we keep the genders evenly distributed by rotating through 4 consecutive actor IDs as the testing set for each fold.

\paragraph{Training.}
We first fine-tuned the unimodal models for each fold on RAVDESS. For both unimodal and multimodal training, we used the Adam optimizer \cite{kingma2014adam} with a constant learning rate of $10^{-3}$. The final accuracy reported is the average accuracy over the 6 folds.  

\paragraph{Benchmarks.}
For this task, we implemented multiple recent multimodal fusion algorithms as our benchmarks. We categorize them into the following: 1) simple feature concatenation followed by fully connected layers based on \cite{ortega2019multimodal} and MCBP \cite{fukui2016multimodal} as two early fusion methods, 2) MMTM \cite{joze2020mmtm} as the state-of-the-art intermediate fusion method, 3) averaging, multiplication are two standard late fusion methods; multiplicative layer \cite{liu2018learn} is a late fusion method that adds a down-weighting factor to CE loss to suppress weaker modalities.

\paragraph{Results.}
Table \ref{table1} presents the accuracy of the proposed method in comparison with the implemented benchmarks. Our network surpasses unimodal baselines by over 10\% verifying the importance of multimodal fusion. Early fusion methods did not exceed standard late fusion benchmarks by a significant number, indicating the challenge of finding cross-modal correlations between the complex video network and the 1D audio model in early stages. As expected, intermediate fusion methods out performed late and early methods as they are able to highlight features while they are developed to identify areas of focus in each modality. Our model outperforms the top performer MMTM by 1.74\% while using 19\% less parameters. Compared to the unimodal models, we only introduced 30K parameters in the fusion module.

\begin{table}[t]
  \centering
  \small
    \begin{tabular}{lccc} 
      Model & Fusion Stage & Accuracy & \#Params\\
      \hline
      3D ResNeXt50 (Vis.) & - & 62.99 & 25.88 M\\
      1D CNN (Aud.) & - & 56.53 & 0.03 M\\
      \hline
      Averaging & Late & 68.82 & 25.92 M \\
      Multiplicative $\beta$=0.3 & Late & 70.35 & 25.92 M \\
      Multiplication & Late & 70.56 & 25.92 M \\
      Concat + FC & Early & 71.04 & 26.87 M\\
      MCBP & Early & 71.32 & 51.03 M\\
      MMTM & Inter. & 73.12 & 31.97 M\\
      \hline
      MSAF & Inter. & \textbf{74.86} & \textbf{25.94 M}\\
      \hline
    \end{tabular}
    \caption{Comparison between multimodal fusion benchmarks and ours for emotion recognition on RAVDESS.}
    \label{table1}
\end{table}

\subsection{Sentiment Analysis}
\paragraph{Data Preparation.}
CMU-MOSI \cite{zadeh2016mosi} and CMU-MOSEI \cite{bagher-zadeh-etal-2018-multimodal} are commonly used datasets for multimodal sentiment analysis. We chose to evaluate our method using CMU-MOSEI since it is the next generation of CMU-MOSI provided by the same authors. The CMU-MOSEI dataset contains 22852 annotated video segments (utterances) from 1000 distinct speakers and 250 topics gathered from online video sharing websites. Each utterance is annotated with a sentiment intensity from [-3, 3]. The train, validation and test set contain 16322, 1871, and 4659 samples respectively. Following recent works, we evaluate: 1) mean absolute error (MAE) and Pearson correlation (Corr) for regression, 2) binary accuracy (Acc-2) and F-score, 3) 7 class accuracy (Acc-7) from -3 to 3. For binary classification, we consider [-3, 0) labels as negative and (0, 3] as positive.

In order to draw comparisons with recent works \cite{sun2019learning,hazarika2020misa}. COVAREP \cite{coverap}, FACET \footnote{https://imotions.com/platform/} and BERT \cite{devlin2018bert} features were selected for audio, visual and text modalities respectively. We acquired aligned COVAREP, FACET features and raw text from the public CMU-MultimodalSDK v1.2.0 \footnote{https://github.com/A2Zadeh/CMU-MultimodalSDK} with a sequence length of 50 words for all modalities. The raw text of individual utterances was passed into a pretrained uncased BERT model (not fine-tuned on CMU-MOSEI) and the final encoder layer output was used for text features. Each utterance was passed in separate epochs to avoid adding additional context. 

\paragraph{Training.}
Since the network is relatively small, the unimodal models were not pretrained. The multimodal network was trained using mean squared error (MSE) loss and the Adam optimizer was used with a constant learning rate of $10^{-3}$. For Acc-7, we rounded the output to the nearest integer and clipped to -3 and 3.

\paragraph{Benchmarks.}
We summarize various multimodal fusion methods proposed for sentiment analysis as follows: 1) DCCA \cite{sun2019multimodal} and ICCN \cite{sun2019learning} use Deep CCA to correlate text with audio-visual features, 2) TFN \cite{zadeh2017tfn} and LMF \cite{8752006lmf} perform outer-products on modalities to create a joint representation, 3) MFM \cite{tsai2018learning} and MISA \cite{hazarika2020misa} separates features into modality-specific and modality-invariant. While MFM optimizes a joint generative-discriminative objective allowing for classification and reconstruction of missing modalities, MISA trains modality-specific and multimodal encoders which are then fused using multi-headed self-attention before prediction.

\paragraph{Results.} 
Table \ref{table2} shows the results of our experiments in comparison with the state-of-the-art and recent works using BERT. Our multimodal model outperforms all unimodal models confirming that audio-visual features improve sentiment analysis. MSAF achieves better or similar performance on all metrics compared to state-of-the-art multimodal methods while using a simpler network architecture. Compared to MISA and others, our training method is simpler, allowing MSAF to be easily applied to existing unimodal networks.

\begin{table}[t]
  \centering
  \small
    \begin{tabular}{lccccc}
      Model & Acc-2 & F-score & MAE & Acc-7 & Corr\\
      \hline
      LSTM (Visual) & 65.9 & 64.7 & 0.829 & 41.3 & 0.248\\
      LSTM (Audio) & 65.7 & 62.7 & 0.821 & 41.2 & 0.263\\
      LSTM (Text) & 83.5 & 83.5 & 0.572 & 51.0 & 0.727\\
      \hline
      DCCA$^\diamond$ & 83.6 & 83.8 & 0.579 & 50.1 & 0.707\\
      TFN$^\diamond$ & 82.6 & 82.1 & 0.593 & 50.2 & 0.700\\
      LMF$^\diamond$ & 82.0 & 82.2 & 0.623 & 48.0 & 0.677\\
      MFM$^\diamond$ & 84.4 & 84.4 & 0.568 & 51.4 & 0.717\\
      ICCN* & 84.2 & 84.2 & 0.565 & 51.6 & 0.713\\
      MISA* & \textbf{85.5} & 85.3 & \textbf{0.555} & 52.2 & \textbf{0.756}\\
      \hline
      \multirow{2}{*}{MSAF} & \textbf{85.5} & \textbf{85.5} & 0.559 & \textbf{52.4} & 0.738\\
      & \textit{0.3} & \textit{0.3} & \textit{0.003} & \textit{0.1} & \textit{0.002} \\
      \hline
    \end{tabular}
    \caption{Comparison between multimodal fusion benchmarks and ours for sentiment analysis on CMU-MOSEI. * from original papers and $^\diamond$ from \cite{sun2019learning}. The standard error over 5 runs are in italics.}
    \label{table2}
\end{table}

\subsection{Action Recognition}
\paragraph{Data Preparation.}
NTU RGB+D \cite{shahroudy2016ntu} is a large-scale human action recognition dataset. It contains 60 action classes and 56,880 video samples associated with 3D skeleton data. Cross-Subject (CS) and Cross-View (CV) are two recommended protocols. CS splits the training set and testing set by the subject IDs, whereas CV splits the samples based on different camera views. Recent methods \cite{yang2020feedback, liu2018recognizing, de2020infrared} have achieved decent CV accuracies; however, CS still remains a more challenging evaluation method based on the reported performance compared to the CV counterpart. We adopt the CS evaluation and split the 40 subjects based on the specified rule. For data preprocessing, video frames are extracted at 32 FPS and we adopt the same data augmentation approach as \cite{joze2020mmtm}. 

\paragraph{Training.}
The Adam optimizer with a base learning rate of $10^{-3}$ and a weight decay of $10^{-4}$ is used. The learning rate is reduced to $10^{-4}$ at epoch 5, where the loss is near saturation in our experiment. 

\paragraph{Benchmarks.}
We summarize the multimodal fusion benchmarks for action recognition based on RGB videos and skeletons as follows: 1) SGM-Net \cite{li2020sgm} proposed a skeleton guidance block to enhance RGB features, 2) CentralNet \cite{vielzeuf2018centralnet} adds a central branch that learns the weighted sum of skeleton and RGB features at various locations, 3) MFAS \cite{Perez-Rua_2019_CVPR} is a generic search algorithm that finds an optimal architecture for a given dataset, 4) PoseMap \cite{liu2018recognizing} uses CNNs to process pose estimation maps and skeletons independently with late fusion for final prediction, 5) MMTM \cite{joze2020mmtm} recalibrates features at different stages achieving state-of-the-art in RGB and skeleton fusion.

\begin{table}[t]
  \centering
  \small
    \begin{tabular}{lcc} 
      Model & RGB Model & Acc. (CS)\\
      \hline
      Inf. ResNet50 (RGB) & - & 83.91 \\
      I3D (RGB) & - & 85.63 \\
      HCN (Skeleton)& - & 85.24 \\
      \hline
      SGM-Net* & - & 89.10 \\
      CentralNet$^\diamond$ & Inf. ResNet50 & 89.36 \\
      MFAS* & Inf. ResNet50 & 90.04 \\
      MMTM* & Inf. ResNet50 & 90.11 \\
      PoseMap* & - & 91.71 \\
      MMTM* & I3D & 91.99 \\
      \hline
      \multirow{1}{*}{MSAF} & \multirow{1}{*}{Inf. ResNet50} & \textbf{90.63}\\
      \multirow{1}{*}{MSAF} & \multirow{1}{*}{I3D} & \textbf{92.24}\\
      \hline
    \end{tabular}
    \caption{Comparison between multimodal fusion benchmarks and ours on the NTU RGB+D Cross-Subject protocol. * from original papers and $^\diamond$ from \cite{joze2020mmtm}. The standard error for Inflated ResNet50 and I3D over 5 runs is 0.04 and 0.03 respectively.}
    \label{table3}
\end{table}

\paragraph{Results.}
Table \ref{table3} reports the accuracy of the proposed MSAF network in comparison with other action recognition models using RGB videos and skeletons. To compare with state-of-the-art intermediate fusion methods, we also evaluate the performance of MSAF applied to Inflated ResNet50 \cite{Baradel_2018_CVPR} and HCN. Our model outperforms all intermediate fusion methods and application-specific models, achieving state-of-the-art performance in RGB+pose action recognition in the NTU RGB+D CS protocol.

\subsection{Ablation Study}
To obtain the configurations used for each application, we conduct the ablation study on all three datasets with the following hyperparameters: the number of channels in a block $C$, attention regularizer $\lambda$ (default value is 0), BlockDropout (with $p=0.2$), and the number of segments $q$ for RNNs (default value is 1). Table \ref{table4} reports the accuracy of the configurations building up to the best configuration. We observe the optimal number of channels in a block, $C$, for each dataset can be derived from $\min{\{ C_1, \cdots, C_{M}\}}/2$ which serves as a good starting point when tuning $C$ for other applications. Hyperparameter $\lambda$ plays an important role in NTU by avoiding over-suppression of features for more complex tasks. BlockDropout is essential to the performance in RAVDESS and MOSEI but not NTU as dropout tends to be more effective on smaller datasets to prevent overfitting. The importance of $q$ is shown in MOSEI, where a sequence length of 5 words was optimal for a sequence length of 50 words. As $q$ increases, the number of MSAF modules increases which increases the number of parameters and thus results in overfitting.

\begin{table}[t]
  \centering
  \small
    \begin{tabular}{lccccc} 
      Dataset & $C$ & $\lambda$ & BlockDropout & $q$ & Acc.\\
      \hline
      \multirow{6}{*}{RAVDESS} & 8, 16 & {} & {} & {} & 71.01 \\
      & 16, 32 & {} & {} & {} & \textbf{72.99} \\
      & 32, 64 & {} & {} & {} & \textbf{73.40} \\
      & 32, 64 & {} & \ding{51} & {} & 72.29 \\
      & 16, 32 & {} & \ding{51} & {} & \textbf{74.86} \\
      & 16, 32 & 0.25 & \ding{51} & {} & 74.37 \\
      \hline
      \multirow{6}{*}{MOSEI} & 8 & {} & {} & {} & 51.6 \\
      & 16 & {} & {} & {} & \textbf{51.7} \\
      & 32 & {} & {} & {} & 51.1 \\
      & 16 & {} & \ding{51} & {} & \textbf{52.3} \\
      & 16 & {} & \ding{51} & 5 & \textbf{52.4} \\
      & 16 & {} & \ding{51} & 10 & 52.2 \\
      \hline
      \multirow{6}{*}{NTU} & 32, 128 & {} & {} & {} & 91.04 \\
      & 64, 256 & {} & {} & {} & \textbf{91.56} \\
      & 126, 512 & {} & {} & {} & 91.05 \\
      & 64, 256 & 0.25 & {} & {} & 92.00 \\
      & 64, 256 & 0.5 & {} & {} & \textbf{92.24} \\
      & 64, 256 & 0.5 & \ding{51} & {} & 92.12 \\
      \hline
    \end{tabular}
    \caption{Ablation study of MSAF module hyperparameters. For CMU-MOSEI, Acc-7 is shown for Acc.}
    \label{table4}
\end{table}

The location of a MSAF module in a multimodal network architecture is an important factor for effective feature fusion. On one hand, placing a MSAF module at an earlier part of the network can help unimodal models learn to correlate raw features of each other. On the other hand, using MSAF to fuse high-level features generates more apparent bias towards specific unimodal patterns as the high-level features are more tailored to the task. To analyze the effect of MSAF in different fusion locations on model performance, we define three positions to place MSAF in our action recognition network (I3D + HCN). In the early location, a MSAF receives the concatenated Conv4 features from the two actors in HCN and the third last Inception layer of I3D. The intermediate location is set to be between the Conv5 layer of HCN and the second last Inception layer of I3D. Finally, the late location is at the last I3D Inception layer and the FC7 layer of HCN. We follow $C = \min{\{ C_1, \cdots, C_{M}\}}/2$ while keeping other parameters the same. We train the multimodal network with different combinations of the above fusion locations and report our results in Table \ref{table5}. 

We observe that the combination of intermediate and late fusion achieves the best result among all seven experiments. Interestingly, all experiments that involve early fusion yield similar performance at around 91.9\%. Further, deploying MSAF in all three locations does not achieve better performance than using only intermediate and late fusion. We believe this is because the low-level features at the early position are still underdeveloped to show enough correlation for effective fusion, which results in sub-optimal performance. In summary, we find that multimodal fusion using MSAF is the most effective when applied to a combination of intermediate and high-level features.

\begin{table}[t]
  \centering
  \small
    \begin{tabular}{lccc} 
      Early & Intermediate & Late & Acc. (CS)\\
      \hline
      \ding{51} & {} & {} & 91.93 \\
      {} & \ding{51} & {} & 92.08 \\
      {} & {} & \ding{51} & \textbf{92.11} \\
      \ding{51} & \ding{51} & {} & 91.81 \\
      \ding{51} & {} & \ding{51} & 91.88 \\
      {} & \ding{51} & \ding{51} & \textbf{92.24} \\
      \ding{51} & \ding{51} & \ding{51} & 91.88 \\
      \hline
    \end{tabular}
    \caption{Ablation study of the placement of MSAF modules in early, intermediate and late feature levels on NTU RGB+D.}
    \label{table5}
\end{table}

\subsection{Parameter and Attention Analysis}
\begin{figure}[h]
\centering
\includegraphics[width=0.8\columnwidth]{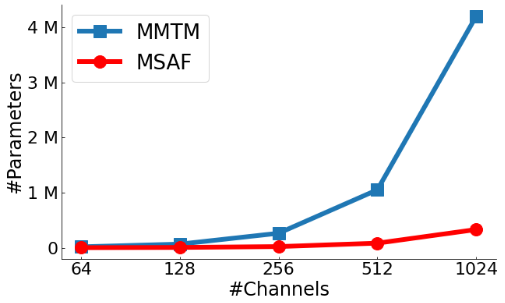} 
\caption{Number of parameters comparison between an MSAF module and an MMTM \cite{joze2020mmtm} module. Each module receives two modalities with the same channel number indicated by the x-axis.}
\label{fig6}
\end{figure}

Reflecting on our objective to design an effective fusion module that is also lightweight, we analyze the number of parameters of the MSAF module. Ideally, the fusion module should introduce minimal parameters to the unimodal networks combined despite the feature map size of the modalities. The split and join steps in MSAF ensure the joint feature space depends on the channel number of the feature blocks instead of the channel of the modalities. Therefore, the number of parameters is significantly reduced. In Figure \ref{fig6}, we compare the number of parameters of MSAF to MMTM \cite{joze2020mmtm}. For both methods, we use two example modalities with shape (4, $\#Channels$, 3, 128, 128) where $\#Channels$ is indicated on the x-axis. The reduction factor is set to 4 for both modules and we set $C = \min{\{ C_1, \cdots, C_{M}\}}/2$ for MSAF. As shown, MSAF utilizes parameters more efficiently, reaching a maximum of 330K parameters. In terms of computational cost, the number of FLOPs for MSAF has a similar trend as the number of parameters. For instance, when $\#Channels$ is 64 and 1024, MSAF has 10.4K and 2.6M FLOPs, whereas MMTM has 131.6K and 33.6M FLOPs respectively.

\begin{figure}[ht]
\centering
\includegraphics[width=0.95\columnwidth]{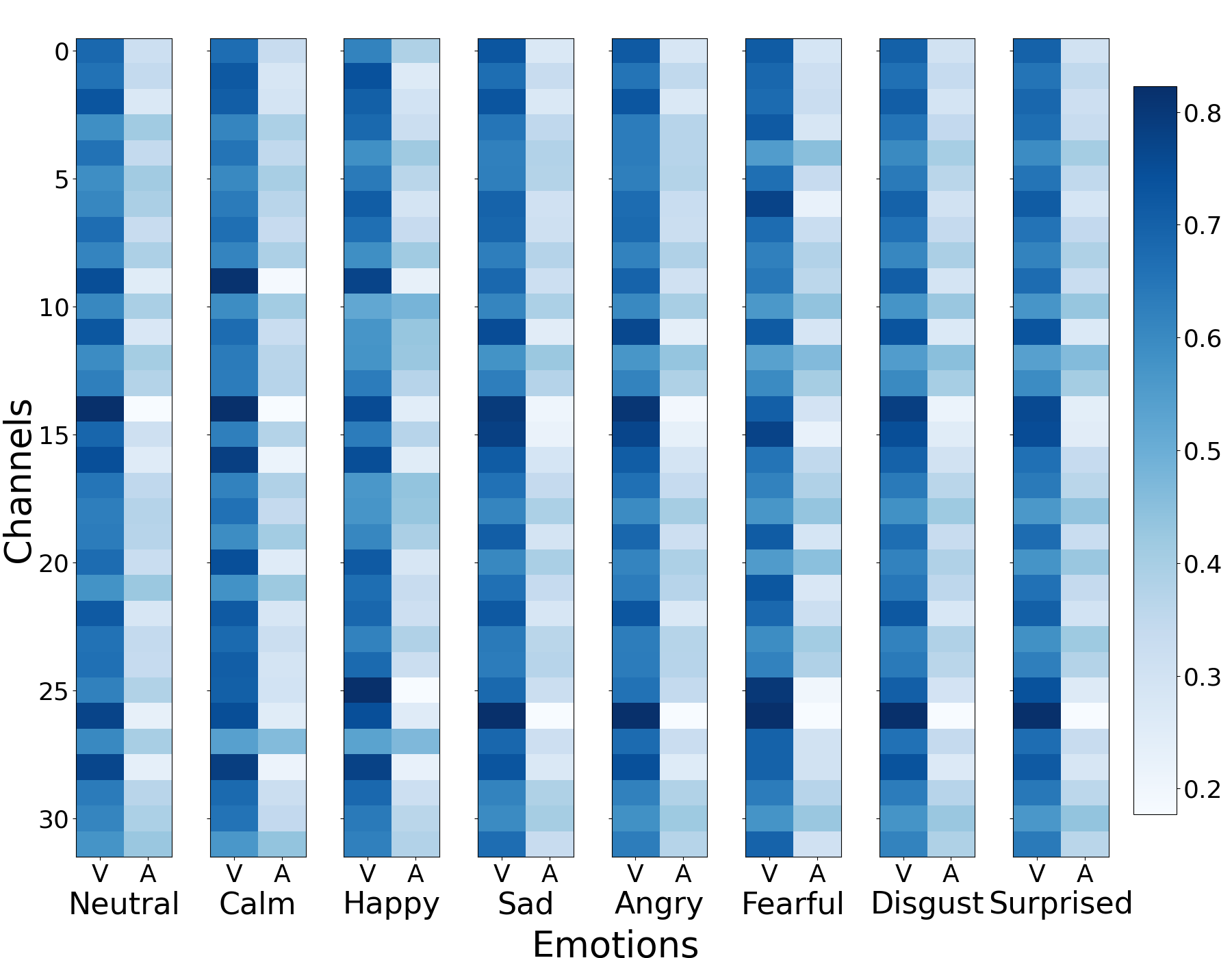} 
\caption{Visualization of attention values from the second MSAF module averaged for each emotion in the RAVDESS dataset and summed modality-wise (V=video, A=audio).}
\label{fig7}
\end{figure}

To further understand the MSAF module and its effectiveness, we analyze the attention signals produced on the RAVDESS dataset. We first compare the attention signals averaged per emotion. Figure \ref{fig7} shows the attention signals from the second MSAF module and sum the attention values for the blocks of the same modality. The video modality has higher attention weights when summed together since it has more blocks and is the stronger modality. However, we observe that for some emotions such as happy, a number of channels in the audio modality have similar weights as the video modality. 
This shows that the MSAF module is able to optimize how the modalities are used together depending on the emotion. 

\begin{figure}[ht]
\centering
\includegraphics[width=0.95\columnwidth]{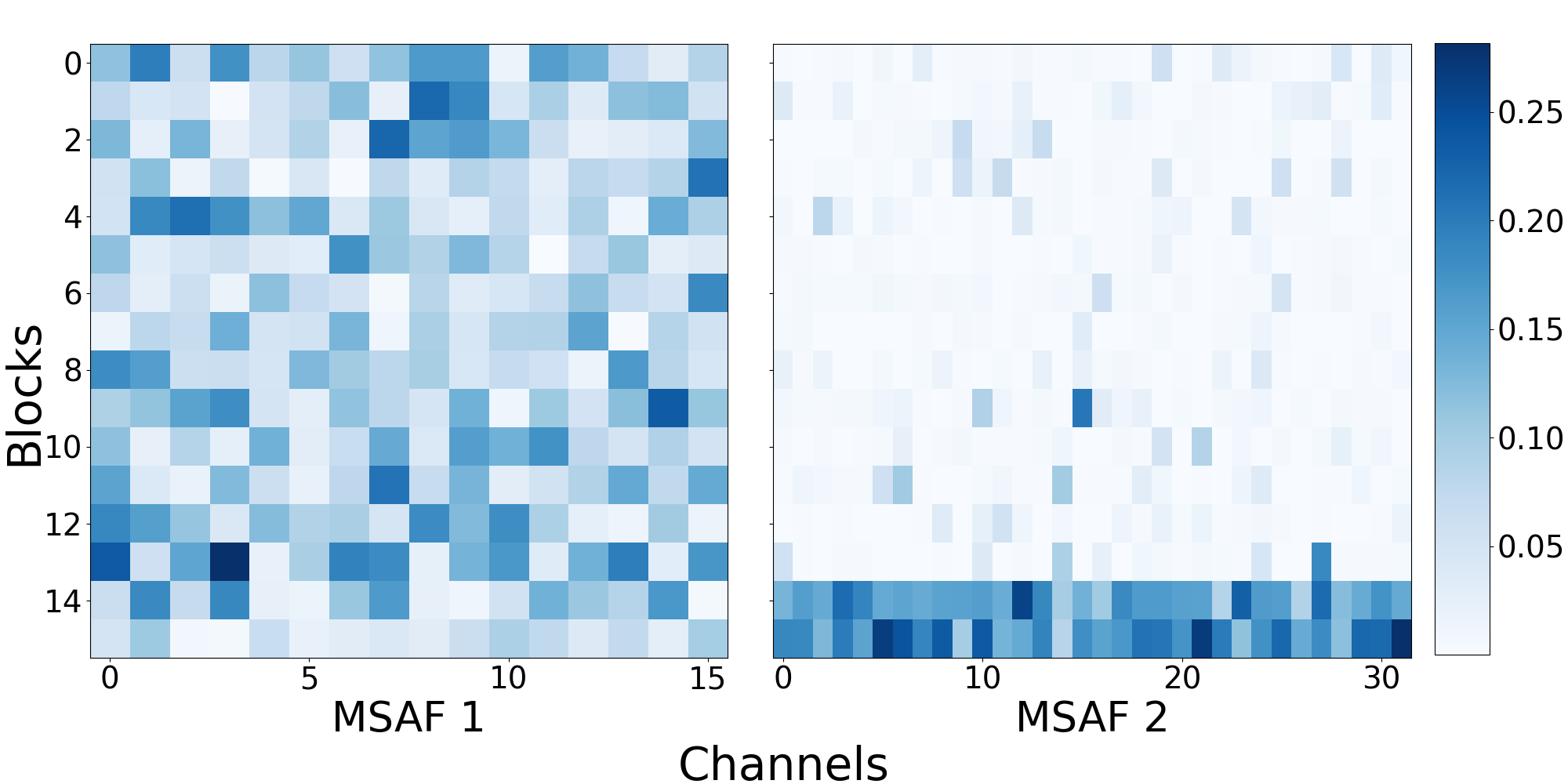} 
\caption{Comparison between attention values of 2 MSAF modules in RAVDESS. Blocks 14-16 belong to the audio modality and part of the video modality is shown due to size.}
\label{fig8}
\end{figure}

Next, we examine the attention signals from the first MSAF module versus the second MSAF module. In Figure \ref{fig8}, the first MSAF module gives blocks of each modality similar levels of attention since the features are lower-level whereas the second MSAF module learns that the audio modality has fewer blocks and gives them higher attention values compared to the video modality blocks. 

\section{Conclusion}
In this work, we present a lightweight multimodal fusion module, MSAF, that learns to exploit the complementary relationships between the modalities and highlight features for optimal multimodal learning. MSAF enables easy deployment of high-performance multimodal models due to its compatibility with CNNs and RNNs. We implement three multimodal networks with MSAF for emotion recognition, sentiment analysis, and action recognition. Our experiments demonstrate the module’s ability to coordinate various types of modalities through competitive evaluation results in all three tasks.

{\small
\bibliographystyle{ieee_fullname}
\bibliography{reference}
}

\end{document}